\documentclass[12pt]{article}
\usepackage[margin=1in]{geometry}                % See geometry.pdf to learn the layout options. There are lots.
\usepackage{setspace}
\usepackage{graphicx}
\usepackage{amsmath}
\usepackage{amssymb}
\usepackage{bm}
\usepackage{url}
\usepackage{epstopdf}
\usepackage{times}
\title{Why Neural Networks Work}
\author{Sayandev Mukherjee and Bernardo A.~Huberman\\ CableLabs\\ \texttt{\{s.mukherjee, b.huberman\}@cablelabs.com}}
\date{}                                           % Activate to display a given date or no date

\begin{document}
\maketitle

\begin{abstract}
We argue that many properties of fully-connected feedforward neural networks (FCNNs), also called multi-layer perceptrons (MLPs), are explainable from the analysis of a single pair of operations, namely a \emph{random projection} into a higher-dimensional space than the input, followed by a \emph{sparsification} operation.  For convenience, we call this pair of successive operations \emph{expand-and-sparsify} following the terminology of Dasgupta.  We show how expand-and-sparsify can explain the observed phenomena that have been discussed in the literature, such as the so-called Lottery Ticket Hypothesis, the surprisingly good performance of randomly-initialized untrained neural networks, the efficacy of Dropout in training and most importantly, the mysterious generalization ability of overparameterized models, first highlighted by Zhang et al.~and subsequently identified even in non-neural network models by Belkin et al.
\end{abstract}

\section{Introduction}
The recent successes of deep learning models in achieving state-of-the-art performance in various tasks has increased the importance of understanding many intriguing aspects of their observed behavior.  Among them are:
\begin{enumerate}
\item Why neural networks perform better than the classical kernel machine ML models like support vector machines and random forests on the same problems;
\item Why neural networks are curiously vulnerable to adversarially crafted inputs that are imperceptibly different (to a human) from inputs which do not result in erroneous output;
\item Why randomly-initialized neural networks, even without training, can perform fairly well;
\item Why vanilla stochastic gradient descent without regularization is usually sufficient to train a neural network;
\item Why a trained neural network can be drastically pruned without loss of performance;
\item Why training a neural network with dropout leads to better performance of the trained model;
\item Why a neural network so overparameterized that it can memorize a training set with arbitrary labels can nevertheless generalize well if trained on a training set with correct labeling, even though it still learns the training set with zero error.  In other words, a neural network seems to \emph{learn} if there is some underlying structure to the training set it is given, and if there is no such structure, then it simply \emph{memorizes} the training set.
\end{enumerate}
The final property in particular, has excited a lot of research ever since it was highlighted by Zhang et al. in~\cite{zbhrv2017, zbhrv2021}.\footnote{Subsequently this property was also reported in non-neural network ML models by Belkin et al.~\cite{bhmm}.}  Explanations for this behavior have included attempts to define new metrics for ``effective complexity" of ML models (see, for example,~\cite{bft}), extrapolations to infinitely-wide neural networks~\cite{cob}, and the use of methods from statistical mechanics to study the dynamics of the interacting neurons of a large neural network~\cite{ryh}.  

In this paper, we will argue that the key to the explanation of this phenomenon lies in something that while somewhat mysterious, it offers excellent heuristic insights -- the properties of a particular mathematical operation that is called \emph{expand-and-sparsify}~\cite{dt}: a \emph{random projection} to a \emph{higher-dimensional} space than the input, followed by a \emph{sparsification} step.

We believe that the components of a successful neural network solution for a given use case are:
\begin{enumerate}
\item Architecture selection: how many layers, how many nodes in these layers, the connections amongst these layers, and the activation functions of these nodes.  In the present work we restrict ourselves to MLPs where all nodes in a given layer are connected to all the nodes in the next higher layer, and only the next higher layer, and all nodes except possibly those in the output layer have ReLU activations~\eqref{eq:relu}.

\item Parameter initialization: The initial values assigned to the parameters of the MLP.  

\item Model training, i.e., parameter optimization via iterative gradient descent on a training dataset.
\end{enumerate}

Nearly all the attention of researchers to date has been focused on model training: its dynamics, both asymptotic (see the Neural Tangent Kernels of Chizat and Bach~\cite{cob}) and non-asymptotic (see the renormalization group approach of Roberts and Yaida~\cite{ryh}), and numerous papers trying to explain why vanilla stochastic gradient descent on mini batches even without regularization yields models with excellent performance on both the training data and new, unseen, test data.

We argue in the present work that this emphasis on training as opposed to architecture and parameter initialization is misplaced.  The reported strong performance of uninitialized neural networks with random parameter selection~\cite{fsm} and the fact that a good initialization allows for considerable pruning of a network without noticeable degradation in performance (part of what is called the Lottery Ticket Hypothesis~\cite{fc}) suggest that the heavy lifting in neural network modeling is done by the architecture and parameter initialization rather than the numerical optimization step of model training.

In the present paper, we will attempt to explain all the above observed phenomena as arising from the interesting and profound properties of the expand-and-sparsify operation performed by the layers of the MLP architecture with suitable parameter initialization.  We argue that in a successful model design, the choice of architecture (number of layers, and number of neurons per layer) and parameter initialization has a zeroth-order impact on model performance even before commencing any training, and that the model training step has only a first-order impact on model performance because it is only a perturbation of parameter values.  As an aside, we interpret the Dropout trick during training~\cite{shkss} as an attempt to compensate for a non-optimal choice of architecture and/or parameter initialization, implying that a better choice of architecture and initialization can allow us to dispense with the need for Dropout during training.

We will begin with a summary of the principal known results about the expand-and-sparsify operation, and their implications for neural network training, almost all of which were recently published by Dasgupta and Tosh in~\cite{dt}.  But first, we need to establish some background. 

\section{Literature Survey}
The usefulness of random features (obtained from random linear projections of input vectors) in machine learning (ML) models for classification and regression has been known for a long time, going back to the design of the Random Forest (RF)~\cite{h, b}.  It was extended to large-scale (at the time) kernel machines in~\cite{rr}.  However, all these random projections were into lower-dimensional spaces than that of the input, so the feature vectors obtained from such random projections were dense vectors of smaller dimension than the input vectors.  Theoretical support for the usefulness of such dimensionality reduction through random projection is provided by the Johnson-Lindenstrauss lemma~\cite{dg}, which states that ``large sets of vectors in a high-dimensional space can be linearly mapped in a space of much lower (but still high) dimension with approximate preservation of distances"~\cite{wjl}.  In other words, input vectors that are close in terms of Euclidean distance will remain close in the lower-dimensional space after random projection.  This fact has been used to infer that random projections (into lower-dimensional, often just two- or three-dimensional, space) implement a form of \emph{locality sensitive hashing} (LSH).

Recently, however, a different form of LSH was discovered in a biological neural network, viz., the olfactory system of the fruit fly~\cite{dsn2017}.  Although it also involves a random projection, the projection is not directly into a low-dimensional space.  Rather, the random projection is into a space of dimension \emph{higher} than the input, after which the projected vector is then sparsified by retaining only the highest few percentile of its values and setting all the rest to zero.  Proofs of the LSH property were given in~\cite[Lem.~1]{dsn2017} and~\cite[Thm.~3]{pv}.\footnote{The latter result, however, is proved for a sparsification yielding the same number of significant entries as the dimension of the input, and relies on an approximation of the Binomial by the Normal distribution whose accuracy was not validated in the paper.}  

The above is an example of a simple expand-and-sparsify operation where the random feature vector is larger in dimension than the input vector, but is also sparse with a constant number of nonzero entries.  A more sophisticated expand-and-sparsify operation would sparsify the entries in the dense higher-dimensional (than the input) vector obtained by random projection by comparing them against a threshold (that is different from one entry to another, and could even be input-dependent) and retaining only those that exceed the threshold.  Note that this sparsification operation leaves us with a number of nonzero entries in the projection that is a random variable and no longer constant.  

This new expand-and-sparsify operation was proposed and analyzed for biological networks in~\cite{dssn2018}, then abstracted and studied in more detail in the context of ML by Dasgupta and Tosh in~\cite{dt}.  Since this more sophisticated expand-and-sparsify operation corresponds to the action of a layer of (artificial) neurons with ReLU activation, the analysis in~\cite{dt} will be the starting point for our discussion in the present work.

\section{The Expand-and-Sparsify Operation}
\subsection{Almost-orthogonality in high dimensions}
Before we review known results for the expand-and-sparsify operation, let us begin by reviewing an even more basic result~\cite[Thm.~3.4]{kk} which provides the heuristic justification for the expansion random projection in expand-and-sparsify: 
%for any $\epsilon \in (0,1)$, 
%\begin{equation}
%	\dim_\epsilon n \geq \frac{2^{n-1}}{\displaystyle{\sum_{k=0}^{\lceil 2 n (1-\epsilon) - 1\rceil}\binom{n}{k}}},
%	\label{eq:kk}
%\end{equation}
%where $\dim_\epsilon n$ is the $\epsilon$-quasiorthogonal dimension of $n$-dimensional Euclidean space, i.e., the size of the largest set of points in the unit sphere\footnote{Recall that the unit sphere is the surface of the unit ball, so all points on the unit sphere have unit Euclidean norm.} such that any two distinct points are ``almost orthogonal" in the sense that the magnitude of their inner product is $\leq\epsilon$.
%
%In other words, 
there exist sets with an exponential (in $n$) number of almost-orthogonal vectors (i.e., vectors whose inner products with one another are all small) in $n$-dimensional Euclidean space.  Moreover, it turns out that such sets are typical in the following sense in high-dimensional Euclidean space~\cite{gtps}: ``With probability close to one an exponentially large number of random vectors are pairwise almost orthogonal and do not span an arbitrarily selected normalized vector if coefficients in linear combinations are not allowed to be arbitrarily large."

\subsection{Random projection with almost-orthogonal vectors}
It follows that one can simply select an exponentially large number $d$ of vectors $\bm{w}_1,\dots,\bm{w}_d$ at random (say, independent and identically distributed, i.i.d., with directionality uniformly distributed on the unit sphere) in a high-dimensional space $\mathbb{R}^n$ and have high confidence that they are all almost orthogonal to one another.  

Consider the $d\times n$ matrix $\bm{W}$ with $\bm{w}_1^\top,\dots,\bm{w}_d^\top$ as its rows, which we shall call the \emph{projection} matrix for brevity.  Given a arbitrary vector $\bm{u} \in \mathbb{R}^n$ in this space (which we will for concreteness call the \emph{input} vector), the \emph{projected} vector $\bm{W} \bm{u}$, i.e., the vector whose entries are the inner products of the random vectors $\bm{w}_1,\dots,\bm{w}_d$ with the input vector $\bm{u}$, comprises the ``coordinates" of the input vector $\bm{u}$ along these ``almost-orthogonal" directions.  

Thus, it is intuitively plausible that the entries of the projected vector can serve as \emph{features} for further processing by an ML model that has been fed with this input vector, i.e., that these entries serve as a \emph{representation} of the input vector that may be more useful to an ML training algorithm than the original input when it comes to training an ML model for some task.  When the dimension $d$ of the projected vector is lower than the dimension $n$ of the input vector, which corresponds to the selection of only a small subset of $d$ almost-orthogonal random vectors to compose the rows of the projection matrix, out of the possible exponentially many (in $n$) rows possible, the above intuition is supported by an approximate distance-preserving property called the Johnson-Lindenstrauss lemma.  This is the traditional method of extracting features via random projection.

\subsection{Expansion and sparsification}
What happens if $d$, the number of rows in the projection matrix (each row being a randomly-selected vector, and the rows all being almost-orthogonal as discussed above), is larger than $n$, the dimension of the input vector?  The projected vector now has dimension $d > n$, so the input dimensionality has been \emph{expanded} by this random projection operation. 

Now, any attempt to develop an ML model whose decisions are understandable, and ideally explainable, to humans will necessarily involve a reduction in the dimensionality of the representation from that of the input vector, $n$.  The traditional method of getting a representation achieves this by restricting $d < n$.  However, if $d > n$, then the reduction of the dimensionality of the representation has to be done separately, in a step that we call \emph{sparsification}.

Sparsification of the projected vector when $d > n$ may be carried out either deterministically or stochastically.  In the deterministic approach, we may simply retain a certain fixed number $k$ ($k < n < d$) out of the $d$ entries in the projected vector, say the $k$ largest values, and drop all the other entries in the projected vector, thereby obtaining a sparse representation with just $k$ significant entries.  This appears to be the sparsification method employed in biological neural networks like the olfactory system of the fruit fly, as modeled by Dasgupta et al.~in~\cite{dsn2017}.

In the stochastic approach, the number of significant entries in the sparse representation constructed out of the projected vector is not fixed to be some $k$, but may vary depending on the representation (and therefore the input) vector.  For example, we may retain only the top, say, 5\% of the entries in the projected vector.  More generally, we may compare the projected vector term by term against a (deterministic or random) \emph{threshold} vector selected in some way, and retain only the entries in the projected vector that are larger than the corresponding entries in the threshold vector.  Note that the choice of threshold vector for comparison will determine the number of significant entries in the resulting sparse representation vector.\footnote{For example, the threshold entries may be selected such that on average, $k$ of the $d$ entries of the projected vector exceed their corresponding thresholds~\cite{dt}.}  This number of significant entries in the sparse representation vector is a random variable that could even be significant compared to the dimension $n$ of the input vector. 

\subsection{ReLU MLP layer as expand-and-sparsify operator}
Consider an $n$-dimensional input $\bm{u} \in \mathbb{R}^n$ to a single layer of a feedforward neural network (MLP) comprising $d > n$ neurons, each being a \emph{Rectified Linear Unit} (ReLU) with the following activation function:
\begin{equation}
	\text{ReLU}(x) = \begin{cases}
		x, & x \geq 0, \\
		0, & x < 0. \label{eq:relu}
	\end{cases}
\end{equation}
Suppose the $i$th neuron has $n$ weights represented by the vector $\bm{w}_i = [w_{i,1},\dots,w_{i,n}]^\top \in \mathbb{R}^n$ and threshold parameter $\tau_i \in \mathbb{R}$ so that its output when given input $\bm{u}$ is $v_i = \text{ReLU}(\bm{w}_i^\top \bm{u} - \tau_i)$, $i=1,\dots,d$.  Thus the output of the MLP layer of $d$ such neurons is the vector
\begin{equation}
	\bm{v} = \begin{bmatrix} v_1 \\ \vdots \\ v_d\end{bmatrix} = \begin{bmatrix}
	\text{ReLU}(\bm{w}_1^\top \bm{u} - \tau_1) \\ 
	\vdots\\ 
	\text{ReLU}(\bm{w}_d^\top \bm{u} - \tau_d)
	\end{bmatrix} = \text{ReLU}\left(\begin{bmatrix}
	\bm{w}_1^\top \bm{u} - \tau_1 \\ 
	\vdots\\ 
	\bm{w}_d^\top \bm{u} - \tau_d
	\end{bmatrix}\right) = \text{ReLU}(\bm{W}\bm{u} - \bm{\tau}),
	\label{eq:relu2}
\end{equation}
where $\bm{W}$ is the $d \times n$ matrix whose rows are $\bm{w}_1^\top,\dots,\bm{w}_d^\top$, and $\bm{\tau} = [\tau_1, \dots, \tau_d]^\top$ is the vector of neuron thresholds.

From~\eqref{eq:relu2}, it is clear that if $\bm{w}_1,\dots,\bm{w}_d$ are i.i.d.~(say, uniformly on the unit sphere in $\mathbb{R}^n$), then $\bm{W}\bm{u}$ is the projected input corresponding to the expansion operation on the input $\bm{u}$ with the random projection matrix $\bm{W}$, and the output $\bm{v}$ of the MLP layer is the result of the expand-and-sparsify operation after stochastic sparsification by comparing every entry of the projected input $\bm{W}\bm{u}$ against a threshold value for that entry and retaining only those entries that are greater than the corresponding threshold.  Of course, in practice $\bm{w}_1,\dots,\bm{w}_d$ are only i.i.d.~at the time they are initialized before gradient descent training begins~\cite{gb, hzrs}, but this observation will nevertheless prove useful, as shown later.

With these preliminaries, let us now review the body of known results about such expand-and-sparsify operations.

\section{Function approximation with single hidden layer}
\subsection{Definitions and notation}
We follow the analysis of Dasgupta and Tosh in~\cite{dt}.  They assume that the sparsification leaves a binary vector, i.e., the activation function of the $d$ neurons in the MLP layer is not ReLU~\eqref{eq:relu} but the unit step function.
%\begin{equation}
%	1_{[0,\infty)}(x) = \begin{cases}
%		1, & x \geq 0, \\
%		0, & x < 0. \label{eq:step}
%	\end{cases}
%\end{equation}
It follows that the output of the MLP layer is not given by~\eqref{eq:relu2} but by the binary $d$-dimensional vector
\begin{equation}
	\bm{z} = [z_1,\dots,z_d]^\top \in \{0,1\}^d, 
	\label{eq:step2}
\end{equation}
where for $j=1,\dots,d$,
\begin{equation}	 
	z_j = \begin{cases}
		1, & \bm{w}_j^\top \bm{u} \geq \tau_j, \\
		0, & \text{otherwise}.
		\end{cases}
		\label{eq:stepj}
\end{equation}
Although we write $\tau_j$ for brevity, each entry $\tau_j$ of the threshold vector $\bm{\tau}$ is actually a function of $\bm{w}_j$, defined as the $100(1 - k/d)$th percentile of the random variable $\bm{w}_j^\top \bm{u}$.
%(computed with respect to the true distribution of the input vector $\bm{u})$, i.e., 
%\begin{equation}
%	\tau_j(\bm{w}_j) = \sup\left\{\theta:\ \mathbb{P}\{\bm{w}_j^\top \bm{u} \geq \theta\} \geq \frac{k}{d}\right\}, \quad j=1,\dots,d.
%	\label{eq:tauj}
%\end{equation}
It follows that the vector $\bm{z}$ in~\eqref{eq:step2} is $k$-sparse \emph{in expectation}, i.e., the mean number of $1$s in $\bm{z}$ is $k$.

\subsection{Mathematical Results for Expand-and-Sparsify}
Consider now the problem of approximating a scalar function $f(\cdot)$ of the input $\bm{u}$.
%, i.e., $f: \mathcal{U} \to \mathbb{R}$, where $\mathcal{U}$ is the space of inputs $\bm{u}$.  
Specifically, we want to approximate $f$ using a two-layer MLP where the single hidden layer of $d$ neurons has output~\eqref{eq:step2}, and the output ``layer" of the MLP is a single linear node.  To be precise, the output of the MLP is a weighted average of the hidden-layer outputs $\bm{z}$:
\begin{equation}
	\hat{f}(\bm{u}) = \frac{w_1' z_1 + \cdots + w_d' z_d}{z_1 + \cdots + z_d},
	\label{eq:mlpout}
\end{equation}
whenever the denominator is positive, where $w_1',\dots,w_d'$ are the weights connecting the single output neuron to the $d$ neurons of the hidden layer.  In particular, the input space $\mathcal{U}$ is divided into regions $\mathcal{U}_j = \{\bm{u} \in \mathcal{U}: \bm{w}_j^\top \bm{u} \geq \tau_j\}$ where $z_j = 1$, $j=1,\dots,d$, and $w_j'$ is defined as the expected value of $f(\cdot)$ over the region $\mathcal{U}_j$.
%\begin{equation}
%	w_j' = \mathbb{E}\left[f(\bm{u})\,|\,\bm{u} \in \mathcal{U}_j\right], \quad j=1,\dots,d.
%	\label{eq:wjprime}
%\end{equation}
%Assume that the function $f(\cdot)$ is $\lambda$-Lipschitz with respect to the Euclidean norm:
%\begin{equation}
%	|f(\bm{u}) - f(\bm{v})| \leq \lambda \|\bm{u} - \bm{v}\|_2 \text{ for all } \bm{u}, \bm{v} \in \mathcal{U}.
%	\label{eq:lip}
%\end{equation}

If all weights of the hidden-layer weight matrix $\bm{W}$ are i.i.d.~zero-mean Gaussian and the distribution of the input $\bm{u}$ to the MLP is supported on an $m$-dimensional ($m < n$) submanifold of $\mathcal{U}$ with some other mild conditions, then for $k \gg \log d$ the function approximation error made by~\eqref{eq:mlpout} is upper bounded by $O((k/d)^{1/m})$~\cite[Thm.~8]{dt}.
%we have the following function approximation result for~\eqref{eq:mlpout}~\cite[Thm.~8]{dt}:
%\begin{equation}
%	\sup_{\bm{u} \in \mathcal{U}} |\hat{f}(\bm{u}) - f(\bm{u})| \leq 4\lambda \left(\frac{k}{c_1 d}\right)^{1/m},
%	\label{eq:iidrp}
%\end{equation}
%for some constant $c_1$.

Note that although this is a remarkable universal approximation result holding for an untrained neural network with random hidden-layer weights selected independent of the input, it does require the sparsity $k$ of the hidden-layer output to be $\gg \log d$.\footnote{In~\cite{dt}, Dasgupta and Tosh say that this is to ensure that every input in the submanifold excites at least one neuron in the hidden layer.}  

Recall that $d$, the number of random vectors selected as rows of the random projection matrix $\bm{W}$, can be exponential in the dimension $n$ of the input $\bm{u}$, so $\log d \sim n$, which means that $k$, the sparsity of the output of the hidden layer of the MLP, could be a significant fraction of $n$, although it is unlikely that we need such a large $k$ for approximation accuracy.

Dasgupta and Tosh also derive a result (see~\cite[Thm.~9]{dt}) that imposes no lower bound on sparsity $k$ of the projected vector $\bm{W}\bm{u}$ when the distribution of the projection weights $\bm{W}$ is mildly dependent on the distribution of the input $\bm{u}$, but that result is for deterministic sparsification, i.e., only the $k$ highest values of $\bm{W}\bm{u}$ are retained.  These assumptions do not describe an MLP, but they do give hope that similar results may be available for the stochastic sparsification that occurs in an MLP.

We will now try to explain several unusual or puzzling phenomena that have been empirically observed in MLPs, some of which have later been observed in other ML models as well.  The canonical use case to keep in mind during our ensuing discussion is that of using a ``deep" MLP for a classification task, say classifying the handwritten digits in the MNIST dataset~\cite{mnist}.

\section{Explanations of Observed Phenomena}
Let us now go over the phenomena listed in the Introduction.
\begin{enumerate}
\item \textit{Why neural networks perform better than the classical kernel machine ML models like support vector machines and random forests on the same problems}: We believe the reason is that the kernel machines deterministically restrict the dimension of the representation to be much smaller than that of the input.  While this may work for a majority of the inputs, a statistically significant fraction of the input space will be better served by a representation with more dimensions, which is possible with the stochastic sparsification inherent in an MLP layer.  In particular, certain input values (especially noisy inputs) may require a significantly denser (less sparse) representation than most of the other input values, in which case the larger number of significant entries in the output of this MLP layer may also be serving to distribute the noise over many more entries and thereby improve the reliability of the representation.  Some indications of this have been observed in the literature (see~\cite[Secs.~4--6]{bmr}).

\item \textit{Why neural networks are curiously vulnerable to adversarially crafted inputs that are imperceptibly different (to a human) from inputs which do not result in erroneous output}: In~\cite[Suppl.~Info.]{dssn2018}, Dasgupta et al.~define and study the probability that two inputs $\bm{u}$ and $\bm{u}'$ that are close in $\mathbb{R}^n$ will yield expand-and-sparsify binary representations $\bm{z}$ and $\bm{z}'$ respectively whose entries are highly correlated.  Although this correlation is high, i.e., $\bm{z}^\top\bm{z}'$ is tightly concentrated (see~\cite[Lem.~4, Suppl.~Info.]{dssn2018}), it is intuitively clear that it should be possible to craft an input $\bm{u'}$ that is very close to a known observed input $\bm{u}$ but whose sparse binary representation $\bm{z'}$ differs from $\bm{z}$, the representation of $\bm{u}$.  An \emph{adversarial input} $\bm{u'}$ is one such input, chosen such that the difference between $\bm{z'}$ and $\bm{z}$ cascades progressively up the higher layers of the MLP until the final output of the MLP, typically a classification of $\bm{u'}$, is different from that of $\bm{u}$. Our intuition would say that such vulnerability to adversarial inputs increases with the number of layers of the MLP.

\item \textit{Why randomly-initialized neural networks, even without training, can perform fairly well}: this is explained as a direct consequence of the high approximation accuracy of~\eqref{eq:mlpout} as discussed above. 

\item \textit{Why vanilla stochastic gradient descent without regularization is usually sufficient to train a neural network}: As remarked in~\cite{cob}, often the weights of the MLP change by very little from their random initial values even after training. This is readily explainable in light of the above phenomenon which shows that the randomly initialized weights are often ``good enough."  If these initial weights require hardly any fine-tuning for even better performance, then it also follows that we need not be concerned with issues such as gradient decay prevention or the curious effectiveness of such a simple unregularized gradient descent optimization method -- they are mostly unnecessary because the expand-and-sparsify operation with a large-enough expansion will deliver good to excellent results even before any numerical optimization is attempted.  The gradient descent iterations during training serve only to perturb these initial values of the parameters\footnote{These perturbations have been analyzed asymptotically by Chizat and Bach in~\cite{cob} and more realistically in great detail in the book-length treatment of Roberts and Yaida~\cite{ryh}.}, and therefore deliver only a first-order impact on model performance, in contrast to the zeroth-order impact of architecture and parameter initialization.

\item \textit{Why a trained neural network can be drastically pruned without loss of performance}: if the expansion of the expand-and-sparsify stage is to a large enough dimension, then there are many possible choices for the sparsification factor that should yield good performance.  It is plausible that aggressive sparsification will lead to many hidden-layer nodes not being activated by any input in the training and/or test set, which allows for their deletion without any effect on performance.\footnote{The observation that a trained neural network can be pruned aggressively without significant impact on performance has been previously noted in the literature and described as part of the Lottery Ticket Hypothesis~\cite{fc}.} However, this intuition also indicates that aggressive pruning will render the MLP more susceptible to adversarial attack.  Moreover, it also implies that the model may underperform on selected test sets comprising natural (unmodified) inputs from regions of the input space that require milder sparsification for optimum processing.  This conjecture should be tested.    

\item \textit{Why training a neural network with dropout leads to better performance of the trained model}: the explanation for this is the same as that of the previous phenomenon, namely that there is a range of choices for sparsification that will all yield trained models that perform well.  Dropout~\cite{shkss} simply corresponds to a further level of stochasticization of the sparsification step, and so it is not surprising that it can do well in scenarios calling for high sparsification in the expand-and-sparsify operation, although another effect of dropout may be to create stochastic ensembles of MLP models with little interaction between them, thereby yielding the performance boost from ensembles~\cite{hss}.  Our intuition implies that the effect of dropout is comparable to increasing the sparsification factor, and this is also a conjecture that should be tested.

\item \textit{Why a neural network so overparameterized that it can memorize a training set with arbitrary labels can nevertheless generalize well if trained on a training set with correct labeling, even though it still learns the training set with zero error}: The most significant consequence of the expand-and-sparsify operation is that ``overparameterization" is a red herring, since a wide range of values of the expansion factor and sparsification factor will yield models with good performance, and a judicious selection of these two factors may even obviate the need for much training of the weights in the corresponding hidden layers. In other words, it is misleading to simply count the number of weights in the hidden layers and compare them to the number of dimensions in the input, or the number of training inputs, and then declare the MLP as a whole to be overparameterized or not as the case may be.  For a training set (think of MNIST) whose examples have been assigned random labels, we know that the expand-and-sparsify operation can approximate even this scrambled label assignment accurately, although the estimation of the linear weights $w_j'$ via backpropagation may require a large number of training steps because of the complexity of the regions $\mathcal{U}_j$ in this case compared to the case with correctly labeled training data.  This increase in required training time over the scrambled-labels dataset has been reported in the literature~\cite{zbhrv2017}.
\end{enumerate}

\section{Next steps}
Much work remains to be done, both in theory and in practice, to close the gap between the tantalizing preliminary results reported in~\cite{dt} and a rigorous body of knowledge that not only explains observed phenomena in neural networks, but can also be used to design better-performing neural networks.

We have made two testable conjectures in connection with our discussion of phenomena 5 and 6 in the previous section.  Our next step should be to test these conjectures.  

Another inference that follows from the expand-and-sparsify methodology is that a wide first hidden layer should be used for better performance of an MLP.  However, the interplay between this wide first hidden layer and the folk wisdom to go ``deep rather than wide" when designing such neural network architectures needs to be examined.  The theoretical reason for going deep was the reduction in complexity required for function approximation, but if, as we have argued above, the expand-and-sparsify operation renders conventional notions of complexity irrelevant, then the wisdom of going deep needs to be scrutinized in depth.  The optimum depth-to-width results derived in~\cite{ryh} should be compared to empirical observations for different expansion and sparsification factors for MLPs of different numbers of layers.

From a theoretical perspective, we suggest as ``best-practice" the choice of expansion and sparsification factors in the expand-and-sparsify model for MLPs with different numbers of total layers.  The result~\cite[Thm.~3]{dt} on the size $d$ of the MLP hidden layer required for a given accuracy of function approximation with a $k$-sparse representation after expand-and-sparsify is only applicable to deterministic sparsification (i.e., retaining the $k$ largest values). Thus, it needs to be generalized both to stochastic sparsification and to adapt to inputs that live on a lower-dimensional subdomain of the $n$-dimensional input space.

%\begin{figure}[htbp]
%\begin{center}
%%\includegraphics[width=\linewidth]{Fig1.pdf}
%\caption{Placeholder for a figure.}
%\label{fig:bandwidth}
%\end{center}
%\end{figure}

\end{document}